\documentclass[runningheads]{llncs}
\usepackage{graphicx}
\usepackage[utf8]{inputenc}
\usepackage{tikz}
\usepackage{float}
\usepackage{hyperref}

\begin{document}
\title{Generating large labeled data sets for laparoscopic image processing tasks using unpaired image-to-image translation}
\titlerunning{Laparoscopic unpaired image-to-image translation}
%
\author{Micha Pfeiffer\inst{1}\and Isabel Funke\inst{1} \and Maria R. Robu\inst{2,3}  \and Sebastian Bodenstedt\inst{1} \and Leon Strenger\inst{1} \and Sandy Engelhardt\inst{4} \and Tobias Roß\inst{5} \and Matthew J. Clarkson\inst{2,3} \and Kurinchi Gurusamy\inst{6} \and Brian R. Davidson\inst{6} \and Lena Maier-Hein\inst{5} \and Carina Riediger\inst{7} \and Thilo Welsch\inst{7} \and Jürgen Weitz\inst{7} \and Stefanie Speidel\inst{1}}
\authorrunning{M. Pfeiffer et al.}
%
\institute{Translational Surgical Oncology, National Center for Tumor Diseases, Dresden, Germany \\
\email{micha.pfeiffer@nct-dresden.de}\\ \and
Wellcome/EPSRC Centre for Interventional \& Surgical Sciences, University College London, UK\\ \and
Centre for Medical Image Computing, University College London, UK\\ \and
Faculty of Computer Science, Mannheim University of Applied Sciences, Germany\\ \and
German Cancer Research Center, Heidelberg, Germany\\ \and
Division of Surgery and Interventional Science, University College London, UK\\ \and
Department for Visceral, Thoracic and Vascular Surgery, University Hospital Dresden, Germany}
\maketitle              
\begin{abstract} In the medical domain, the lack of large training data sets and benchmarks is often a limiting factor for training deep neural networks.
In contrast to expensive manual labeling, computer simulations can generate large and fully labeled data sets with a minimum of manual effort. However, models that are trained on simulated data usually do not translate well to real scenarios.
To bridge the domain gap between simulated and real laparoscopic images, we exploit recent advances in unpaired image-to-image translation.
We extent an image-to-image translation method to generate a diverse multitude of realistically looking synthetic images based on images from a simple laparoscopy simulation.
By incorporating means to ensure that the image content is preserved during the translation process, we ensure that the labels given for the simulated images remain valid for their realistically looking translations.
This way, we are able to generate a large, fully labeled synthetic data set of laparoscopic images with realistic appearance.
We show that this data set can be used to train models for the task of liver segmentation of laparoscopic images. We achieve average dice scores of up to 0.89 in some patients without manually labeling a single laparoscopic image and show that using our synthetic data to pre-train models can greatly improve their performance. The synthetic data set will be made publicly available, fully labeled with segmentation maps, depth maps, normal maps, and positions of tools and camera (\url{http://opencas.dkfz.de/image2image}).


\end{abstract}

\begin{figure}
    \begin{tikzpicture}[font=\boldmath]
    \node (top) at (0,0) {\includegraphics[width=\textwidth]{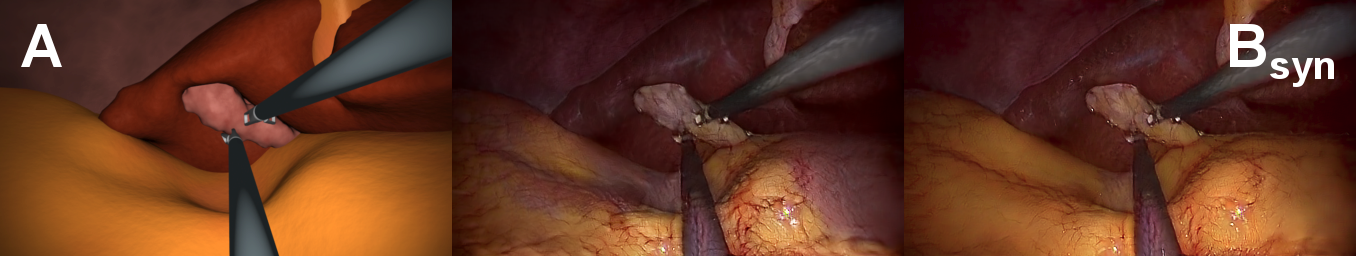}};
    \node (bottom) at (0,-2.3) {\includegraphics[width=\textwidth]{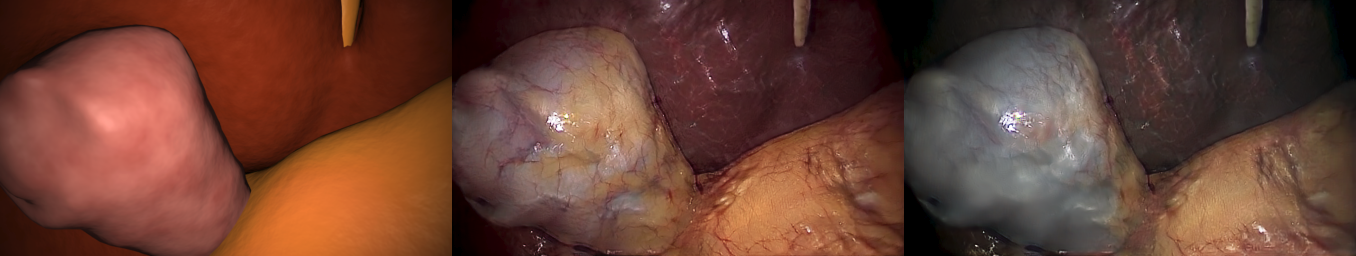}};
    \end{tikzpicture}
    \caption{Images from simple laparoscopic computer simulation (domain $A$, first column) translated to look like real laparoscopic video frames (synthetic $B_{syn}$, second and third column) using various styles.
    During the unpaired training process, a multi-scale structural similarity loss ensures that structures remain similar. This enables us to use the generated images along with labels from domain $A$ as training data for various tasks.}
    \label{fig:LaparoscopicI2I}
\end{figure}

\section{Introduction}

With the increase in computing power, there is an obvious trend towards training larger and deeper networks. However, in the medical domain, the lack of large data sets is a strong limiting factor \cite{maier2017surgical}. The difficulty of recording real patient data in an operating room, legal restrictions on sharing and the great expense of manual labeling by experts make it near impossible to generate large training benchmarks. This work focuses on the example of the segmentation of laparoscopic videos, where deep networks can achieve high accuracies, but sometimes fail to generalize to new patients due to the lack of more labeled data \cite{Gibson2017}.
A solution to this problem could be the usage of synthetic training data. In computer simulations, large amounts of fully labeled data can be created automatically. The main issue here is that models trained on synthetic data usually do not generalize well to real data, due to the \emph{domain gap} between the two.

Instead, we propose to use \emph{image-to-image translation} techniques to translate images from the domain of simulated images in which labels are known (domain $A$), to the domain of real images in which we want to train our model (domain $B$). Recent advances in image translation make it possible to do this even if the data is unpaired, i.e. no direct mapping between samples in one domain to samples in the other domain exists \cite{CycleGAN}. Additionally, \emph{multi-modal} image-to-image translation \cite{MUNIT,DRIT} enables us to control the style of the translation result, which can be utilized to increase the diversity in the final data set.
In the present work, domain $A$ consists of images from very simple laparoscopic 3D computer simulations while domain $B$ is the domain of images from real laparoscopy video feeds.
In order to use the translated data for training, care must be taken that a) the translated images look realistic enough to bridge the domain gap and b) the labels remain valid. This is especially difficult in laparoscopic images, since image content can change drastically between different viewpoints and between patients. To achieve our goal, we build up on several methods:

\subsubsection{Unpaired translation}
The CycleGAN \cite{CycleGAN} has made it possible to translate images between two unpaired domains by usage of a cycle consistency loss and adversarial losses. A generator network $G_B$ translates images from $A$ to $B$ which a discriminator network $D_B$ tries to differentiate from real images in $B$. At the same time, generator $G_A$ and $D_A$ use the same method to translate images from $B$ to $A$. The cycle consistency states that an image $a$ translated to $B$ and back to $A$ must match the original image, i.e. $a = G_A(G_B(a))$ (and symmetrically for an image $b$). This method can only learn a one-to-one mapping (uni-modal), meaning each input image will generate exactly one output.

\subsubsection{Multi-Modal translation}
The key idea behind multi-modal image translation is the separation of an image's \emph{content} from its \emph{style}. The assumption is that the content between domains remains the same, while the style is domain-specific (texture, lighting). An encoder $E_A$ first extracts a \emph{style-code $s_a$} and a \emph{content-code $c_a$} from the source image and a generator $G_B$ then uses this content-code together with a style-code $s_b$ \emph{from the target domain} to create the image $b'$ in the target domain \cite{MUNIT,DRIT}. The opposite direction works analogously. A cycle loss and various reconstruction losses bind the networks together.


\subsubsection{Label-preserving translation}
SPIGAN \cite{SPIGAN} proposes to train an additional network which tries to predict the depth map from the translated image, arguing that this preserves image structure. In our experience, this bears the risk of co-adaptation between the networks. AugGAN \cite{AugGAN} and GANTruth \cite{GANTruth} bind the generators to the image structure via weight-sharing with segmentation networks. However, AugGAN requires segmentation labels to be known for both domains and GANTruth requires a pre-trained segmentation network in the target domain. Our goal is to not use labels during the translation process, simplifying the training procedure.


\subsubsection{Contribution}
In this work, we show how both the goal of realism as well as the preservation of label accuracy during translation can be achieved.
First, we build an extension to the MUNIT framework which is asymmetrical and does not require the simulated domain to have multiple styles, speeding up the process of creating the simulated data.
Next, we incorporate an additional \emph{multi-scale structural similarity loss} \cite{MS-SSIM} and show that it helps to preserve image content and structure despite large changes in camera viewpoint. Additionally, we show how the addition of noise in the encoders can help avoid \emph{mode collapse} - where multiple images map to a similar output - and steganography.
To validate the approach, we show that pre-training a segmentation model on the synthetic data can increase segmentation scores.

As part of this work, we translate 100 \nolinebreak000 images to domain $B$ (see Fig. \ref{fig:LaparoscopicI2I}). This data set, fully labeled with segmentation maps, depth maps and further labels as well as the code will be publicly available\footnote{Data set and code available at: \url{http://opencas.dkfz.de/image2image/}}, with possible applications ranking from pre-training to benchmarking.

\section{Methods}

\begin{figure}
    \centering
    \includegraphics[width=\textwidth]{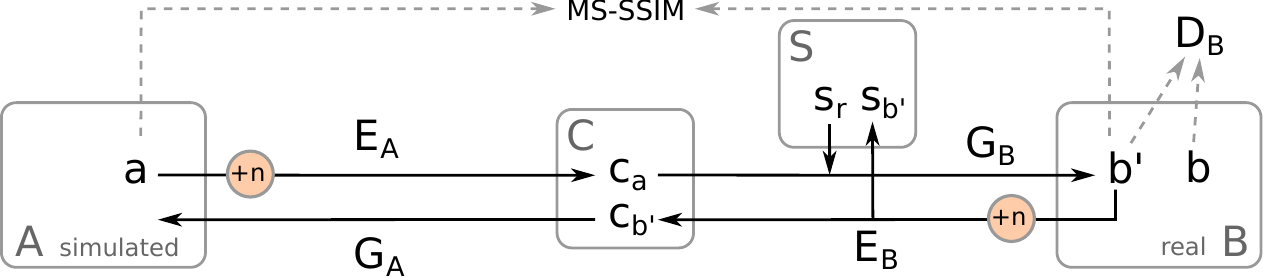}
    \caption{Architecture based on MUNIT \cite{MUNIT}. Image $a$ randomly drawn from $A$ is translated to $B$ and back to $A$, where a cycle loss ensures that $a$ is reconstructed correctly. The same is done in the opposite direction for images drawn from $B$. Various reconstruction losses ensure that the generators and encoders work as expected (please see \cite{MUNIT} for more details). During the translation process, images from $A$ are encoded to a latent code $c_a$, while images from $B$ are split into two latent codes: content $c_b$ and style $s_b$. Unlike MUNIT, we do not have a style in $A$, which simplifies the creation of the rendered images. Furthermore, we add noise to all encoders to prevent the hiding of information and add the MS-SIM loss between source images and their translations.}
    \label{fig:Training}
\end{figure}

Unpaired multi-modal image-to-image translations can output convincing results, but have mostly been tested on scenarios where the content stays similar in all images across both domains (such as faces to faces or mountains to mountains) \cite{MUNIT,DRIT}.
In laparoscopy, viewpoints can change and structures - such as the gallbladder or abdominal wall - move into and out of the view. Incorporating this into our data set is necessary as we want it to be very diverse, however, the mismatch in domain distributions can lead to many wrongly added details, such as a gallbladder where there should only be liver and fat tissue replacing liver tissue.
The following describes our extensions to the MUNIT architecture which enable us to deal with these issues, namely adding a structure-preserving loss, simplifying the encoder $E_A$ and using noise to avoid co-adaptation of the networks.
The resulting training process is outlined in Fig. \ref{fig:Training}.

\subsection{Architecture}

\emph{Multi-Scale Structural Similarity (MS-SSIM) loss}:
Unpaired translation networks often invent details in their output. This is likely due to two reasons: 1) Some structures and some viewpoints occur more in one of the two domains than in the other. For example, domain $A$ contains more close-ups of the liver due to the random placement of the camera. The discriminator $D_B$ will discourage these images, resulting in the generator $G_B$ inventing structures like an additional gallbladder. 2) Generative models are susceptible to mode collapse. We add a multi-scale structural similarity \cite{MS-SSIM} loss between an image $a$ and its translation $G_B(a)$ (and similarly in the other direction). The loss works on the image brightness (average over the channels) which ensures that brighter regions (such as the gallbladder) remain brighter and darker regions remain dark while at the same time not penalizing style-dependent changes in hue.

\emph{Noise against steganography}:
GANs have shown to be very effective hiding information in their output images \cite{Steganography}. Since the generators $G_A$ and $G_B$ are trained jointly to fulfill the cycle consistency, $G_A$ learns to hide details of the image $b$ in its translation which are useful for $G_B$. This is problematic when giving $G_B$ a real image to translate, since these details are not present in this case. To circumvent this effect, we add Gaussian noise to the input of each translation network.

\emph{Asymmetrical style}:
One of our aims is to reduce the amount of manual work required to generate data. In this spirit, we want to translate from a simple and easy to set up domain $A$ to a very complex domain $B$ and let the computer do the bulk of the work automatically.
We remove the part of encoder $E_A$ which extracts the style and the style-injection from $G_A$. As a result, our setup becomes asymmetrical and we do not need to worry about creating multiple textures or lighting styles in the simulated domain $A$, simplifying the simulation process.
During training, both the style extracted by $E_B$ as well as randomly drawn style vectors are used when translating from $A$ to $B$. In this way, the network can later translate images either using a random style or the style taken from a real image.

\subsection{Translation data}
\begin{figure}
    \centering
    \includegraphics[width=0.8\textwidth]{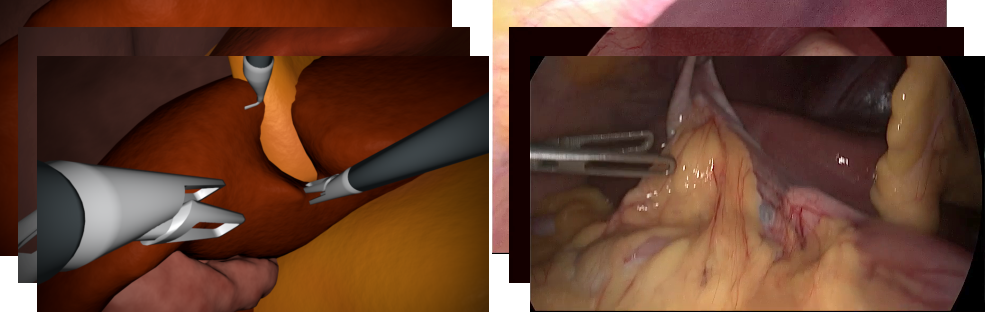}
    \caption{Sample images from the two domains. Both contain similar objects, but no pairing information is known, and the distribution of content does not necessarily match.}
    \label{fig:Domains}
\end{figure}

To train our translation networks, we use two unpaired data sets, which both contain images with livers, gallbladders, tools, fat and abdominal wall (see Fig. \ref{fig:Domains}).

\emph{Rendered data set - Domain $A$}:
We create six synthetic laparoscopic 3D-scenes using the liver and gallbladder surface meshes extracted from CT scans of six patients (3D-IRCADb 01 data set, IRCAD, France). We add meshes which represent fat tissue, ligament and the inflated abdominal wall. Each tissue type is assigned a distinctive texture with small random details. We randomly place the camera together with a light source (representing the laparoscope) and tools. In this way, we render 2000 images from random perspectives for each patient, resulting in 12 \nolinebreak000 synthetic images.
To increase the diversity in our translated results, we repeat the process for four additional patients where no gallbladder is present, resulting in scenes similar to liver staging procedures. The images from all ten patients together make up our extended rendered data set $A^+$.

\emph{Real data set - Domain $B$}:
The real images are taken from 80 videos of the Cholec80 data set (videos of 80 laparoscopic cholecystectomies) \cite{Cholec80}. We first identify parts of the videos in which the gallbladder is still intact and then extract frames at five frames per second. We separate the resulting images into a training data set $B_{tr}$ (75 patients, roughly 74 \nolinebreak000 images) and a segmentation data set $B_v$ (5 patients). We manually segment the liver in 196 images of $B_v$ (at a rate of one frame every five seconds).

\subsection{Experiments}
We train the translation networks for 375 \nolinebreak000 iterations. Afterwards, we translate all images from $A^+$, using five randomly drawn style vectors for each image, resulting in 100 \nolinebreak000 images which we call the synthetic data set $B_{syn}$.

Evaluating the image quality quantitatively is difficult. Instead, we validate the usefulness of the synthetic data set by using it as training data for a segmentation task:
As a baseline, we first train a TernausNet-11 \cite{TernausNet} on the real Cholec80 validation data set $B_v$ in a leave-one-patient-out cross-validation (five models trained, each time one patient is left out of the training data to be used for testing).
We then train the same network only on the synthetic data $B_{syn}$ and validate it on all five patients in $B_v$.
Furthermore, we test how the performance changes if the network which is already trained on $B_{syn}$ is fine-tuned on the real data in the same cross-validation as before.
The experiments are repeated for a TernausNet which has previously been pre-trained on the ImageNet data set \cite{ImageNet}.

To see how our synthetic data helps in the adaptation to a wider diversity of images, we evaluate the pre-initialized TernausNets on images from 13 liver staging sequences, in which a total of roughly 2000 images are segmented \cite{Gibson2017}.

\section{Results}

\begin{figure}[h]
\begin{tikzpicture}[font=\boldmath]
    \centering
    \node[anchor=south west,inner sep=0] (ba) at (0,0) {\includegraphics[width=\textwidth]{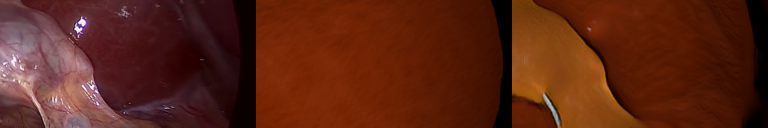}};
    \begin{scope}[x={(ba.south east)},y={(ba.north west)}]
        \node [text=white,ultra thick] at (0.3,0.82){\Large $b$};
        \node [text=white,ultra thick] at (0.6,0.82){\Large $G_A(b)$};
        \node [text=white,ultra thick] at (0.93,0.82){\Large $G'_A(b)$};
        \draw[white,very thick,->] (0.28,0.2) -- (0.36,0.2);
        \draw[white,very thick,->] (0.28,0.1) -- (0.7,0.1);
    \end{scope}
\end{tikzpicture}
\begin{tikzpicture}[font=\boldmath]
    \centering
    \node[anchor=south west,inner sep=0] (ba) at (0,0) {\includegraphics[width=\textwidth]{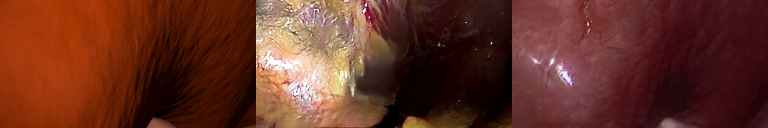}};
    \begin{scope}[x={(ba.south east)},y={(ba.north west)}]
        \node [text=white,ultra thick] at (0.3,0.82){\Large $a$};
        \node [text=white,ultra thick] at (0.6,0.82){\Large $G_B(a)$};
        \node [text=white,ultra thick] at (0.93,0.82){\Large $G'_B(a)$};
        \draw[white,very thick,->] (0.28,0.2) -- (0.36,0.2);
        \draw[white,very thick,->] (0.28,0.1) -- (0.7,0.1);
    \end{scope}
\end{tikzpicture}
    \caption{Qualitative results for the MS-SSIM loss.
    During translation of images $b$ and $a$, the networks tend to remove ($G_A(b)$) or add ($G_B(a)$) detail. In contrast, networks $G'_A$ and $G'_B$, which are trained with an MS-SSIM loss, preserve structures in both directions.
    }
    \label{fig:MS-SSIM}
\end{figure}
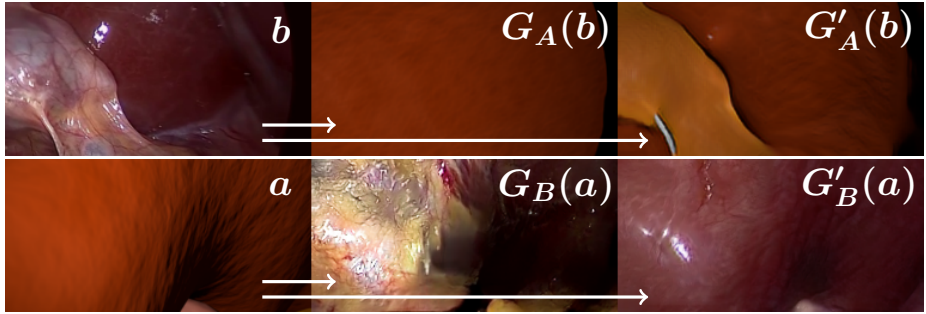

Using the MS-SSIM loss can greatly improve the preservation of image structure, as shown qualitatively in Fig. \ref{fig:MS-SSIM} and helps in the correct usage of textures: The correct assignment of texture to the various organs can be clearly seen and close-up shots of the liver surface result in highly detailed liver texture translations (more translation results in the supplementary materials).

{\renewcommand{\arraystretch}{1.2}
\begin{table}[h]
\caption{Median dice scores for $B_{v}$ (Patients 75 to 80 from Cholec80 data) and for the 13 staging procedures. In all cases where $B_{v}$ is part of the training data, the reported results are from a leave-one-patient-out cross-validation (except for the staging procedures, where all five patients were used). In patient P78 most of the visible liver region is covered in ligament and fat tissue. Median scores for the 13 staging procedures increase considerably by using the synthetic data $B_{syn}$ for pre-training. An additional improvement is achieved by pre-training on the ImageNet data $I$.} 
\begin{tabular}{ p{3cm} p{1cm} p{1cm} p{1cm} p{1cm} p{1.5cm} p{3cm} }
 Training data & P76 & P77 & P78 & P79 & P80 & Staging Procedures\\ \hline
\rule{0pt}{3ex}$B_v$ & 0.50 & 0.68 & 0.42 & 0.52 & 0.56 \\
$B_{syn}$ & 0.73 & 0.70 & 0.13 & 0.74 & 0.76 \\
$B_{syn}$ + $B_v$ & 0.74 & 0.72 & 0.40 & 0.64 & 0.61 \\
\rule{0pt}{3ex}$I$ + $B_v$ & 0.80 & 0.81 & 0.48 & 0.86 & 0.83 & \hfil0.25\\
$I$ + $B_{syn}$ & {0.89} & {0.80} & {0.12} & {0.80} & {0.85} & \hfil0.61 \\
$I$ + $B_{syn}$ + $B_v$ & 0.92 & 0.83 & 0.64 & 0.89 & 0.91 & \hfil0.77\\
\end{tabular}
\label{table:ResultsCholec80}
\end{table}
}
Training on our synthetic data shows considerable improvements over training only on the real data (Table \ref{table:ResultsCholec80}). When using the synthetic data for pre-training, the median dice score improved by an average of 16 percent (no ImageNet pre-training) and 11 percent (with ImageNet pre-training).


When the network was tested on the 13 staging procedures \cite{Gibson2017} containing data that had not been seen at all during training, the mean dice score using only real training data $B_{v}$ was 0.25, and improved to 0.77 when the network was pre-trained with the synthetic data $B_{syn}$.


\section{Discussion}

In this work, we have shown that consistent translation results can be achieved despite having a large change in content and viewpoints.

The translated results alone can be used to achieve reasonably good scores on a segmentation task without labeling a single image. When pre-training a network with our synthetic data, we can demonstrate an increase in performance, compared with only using real data. We also show that the training data can help a network in generalizing to new situations.

Unpaired image-to-image translation is proving to be a very powerful tool in the generation of training data. Since the domain of surgical data science still mostly lacks large benchmarks and open data sets, it could greatly benefit from further development in this field.

\bibliographystyle{splncs04}
\bibliography{bibliography}
%




\end{document}